\documentclass[conference]{IEEEtran}
\IEEEoverridecommandlockouts
\usepackage[hidelinks]{hyperref} 
\usepackage[utf8]{inputenc} 
\usepackage{array,booktabs,ragged2e}
\newcolumntype{R}[1]{>{\RaggedLeft\arraybackslash}p{#1}}
\usepackage{color, colortbl}
\usepackage{multirow}
\definecolor{Gray}{gray}{0.9}
\usepackage[flushleft]{threeparttable}
\usepackage{footnote}
\usepackage{amssymb}
\usepackage{pifont}
\usepackage{graphics}
\usepackage{nicefrac}
\usepackage{lscape}
\usepackage[nottoc,numbib]{tocbibind}
\usepackage[T1]{fontenc} 
\usepackage{mathpazo} 
\usepackage{amsmath}
\usepackage{booktabs}
\usepackage{graphicx}
\usepackage{caption}
\usepackage[labelformat=simple]{subcaption}
\usepackage{bigdelim}
\usepackage{multirow}
\usepackage{adjustbox}
\usepackage{threeparttable}
\usepackage{tabularx}
\newcolumntype{Y}{>{\centering\arraybackslash}X}
\usepackage[ruled, lined, linesnumbered, commentsnumbered, noend]{algorithm2e}
\usepackage[acronym, toc, section=section]{glossaries}
\usepackage{commath} 

\newacronym{ai}{AI}{Artificial Intelligence}
\newacronym{ml}{ML}{Machine Learning}
\newacronym{cv}{CV}{Cross Validation}
\newacronym{fml}{FML}{Fair Machine Learning}
\newacronym{acc}{ACC}{Accuracy}
\newacronym{bacc}{BACC}{Balanced Accuracy}
\newacronym{mcc}{MCC}{Matthews correlation coefficient}
\newacronym{norm_mcc}{NORM\_MCC}{Normalized Matthews correlation coefficient}
\newacronym{dor}{DOR}{Diagnostic Odds Ratio}
\newacronym{fp}{FP}{False Positives}
\newacronym{fpr}{FPR}{False Positive Rates}
\newacronym{fn}{FN}{False Negatives}
\newacronym{fnr}{FNR}{False Negative Rates}
\newacronym{tp}{TP}{True Positives}
\newacronym{tpr}{TPR}{True Positive Rates}
\newacronym{tn}{TN}{True Negatives}
\newacronym{tnr}{TNR}{True Negative Rates}
\newacronym{fdr}{FDR}{False Discovery Rate}
\newacronym{ppv}{PPV}{Positive Predictive Value}
\newacronym{for}{FOR}{False Omission Rate}
\newacronym{ppvd}{PPVD}{Positive Predictive Value Difference}
\newacronym{ford}{FORD}{False Omission Rate Difference}
\newacronym{spd}{SPD}{Statistical Parity Difference}
\newacronym{aod}{AOD}{Average Odds Difference}
\newacronym{eod}{EOD}{Equal Opportunity Difference}
\newacronym{cns}{CNS}{Consistency}
\newacronym{gei}{GEI}{Generalized Entropy Index}
\newacronym{ti}{TI}{Theil Index}
\newacronym{compas}{COMPAS}{Correctional Offender Management Profiling for Alternative Sanctions}
\newacronym{gc}{GC}{German Credit}

\newacronym{bm}{BM}{bias mitigation}
\newacronym{rw}{RW}{Reweighing}
\newacronym{lfr}{LFR}{Learning Fair Representations}
\newacronym{ad}{AD}{Adversarial Debiasing}
\newacronym{egr}{EGR}{Exponentiated Gradient Reduction}
\newacronym{roc}{ROC}{Reject Option Classifier}
\newacronym{ceo}{CEO}{Calibrated Equalized Odds}
\newacronym{pr}{PR}{Prejudice Remover}
\newacronym{dir}{DIR}{Disparate Impact Remover}
\newacronym{lr}{LR}{Logistic Regression}
\newacronym{rf}{RF}{Random Forest}
\newacronym{gb}{GB}{Gradient Boosting}
\newacronym{svm}{SVM}{Support Vector Machine}
\newacronym{nb}{NB}{Naive Bayes}
\newacronym{tabtrans}{TabTrans}{TabTransformer}
\newacronym{nn}{NN}{Neural Networks}
\newacronym{dt}{DT}{Decision Tree}
\def\BibTeX{{\rm B\kern-.05em{\sc i\kern-.025em b}\kern-.08em
    T\kern-.1667em\lower.7ex\hbox{E}\kern-.125emX}}

\begin{document}
\setlength{\abovedisplayskip}{1pt} 
\setlength{\belowdisplayskip}{1pt} 

\setlength{\textfloatsep}{1pt} 
\setlength{\floatsep}{1pt} 

\bstctlcite{IEEEexample:BSTcontrol} 

\title{
FairGridSearch: A Framework to Compare Fairness-Enhancing Models\\
\thanks{
The~pro\_digital European Digital Innovation Hub (EDIH) at Technische Hochschule Wildau received co-funding from the European Union’s DIGITAL EUROPE Programme research and innovation programme grant agreement No. 101083754. Preprint of 979-8-3503-0918-8/23/\$31.00 ©2023 IEEE.  
The published version is available at https://doi.org/10.1109/WI-IAT59888.2023.00064
 }
 }

\author{\IEEEauthorblockN{1\textsuperscript{st} Shih-Chi Ma}
\IEEEauthorblockA{\small
\textit{School of Business and Economics} \\
\textit{Humboldt-Universität zu Berlin}\\
Berlin, Germany\\
\textit{EDIH pro\_digital}\\
\textit{Technische Hochschule Wildau}\\
Wildau, Germany \\
shih-chi.ma@th-wildau.de
}
\and
\IEEEauthorblockN{2\textsuperscript{nd} Tatiana Ermakova}
\IEEEauthorblockA{\small
\textit{School of Computing, Communication and Business} \\
\textit{Hochschule für Technik und Wirtschaft Berlin}\\
Berlin, Germany \\
tatiana.ermakova@htw-berlin.de}
\and
\IEEEauthorblockN{3\textsuperscript{rd} Benjamin Fabian}
\IEEEauthorblockA{\small
\textit{EDIH pro\_digital}\\
\textit{Technische Hochschule Wildau}\\
Wildau, Germany \\
\textit{School of Business and Economics} \\
\textit{Humboldt-Universität zu Berlin}\\
Berlin, Germany \\
benjamin.fabian@th-wildau.de}
}
\maketitle

\thispagestyle{plain} 
\pagestyle{plain} 
\begin{abstract}
  Machine learning models are increasingly used in critical decision-making applications. However, these models are susceptible to replicating or even amplifying bias present in real-world data. While there are various bias mitigation methods and base estimators in the literature, selecting the optimal model for a specific application remains challenging.
  This paper focuses on binary classification and proposes FairGridSearch, a novel framework for comparing fairness-enhancing models. FairGridSearch enables experimentation with different model parameter combinations and recommends the best one. 
  The study applies FairGridSearch to three popular datasets (Adult, COMPAS, and German Credit) and analyzes the impacts of metric selection, base estimator choice, and classification threshold on model fairness.
  The results highlight the significance of selecting appropriate accuracy and fairness metrics for model evaluation. Additionally, different base estimators and classification threshold values affect the effectiveness of bias mitigation methods and fairness stability respectively, but the effects are not consistent across all datasets.
  Based on these findings, future research on fairness in machine learning should consider a broader range of factors when building fair models, going beyond bias mitigation methods alone.
\end{abstract}


\vspace{-3pt}
\section{Introduction}
\label{sec:introduction}
\acrfull{ml} models are increasingly utilized in critical decision-making applications, such as workforce recruiting \cite{zhao_learning_2018, buyl_tackling_2022}, justice risk assessments \cite{angwin_machine_2016, tolan_why_2019}, and credit risk prediction \cite{kozodoi_fairness_2022, kumar_equalizing_2022}. Even though \acrshort{ml} algorithms are not intentionally designed to incorporate bias, studies have shown that \acrshort{ml} models not only reproduce existing biases in the training data \cite{bolukbasi_man_2016} but also amplify them \cite{zhao_men_2017, foulds_intersectional_2019, hall_systematic_2022}. 
Concerns about algorithmic fairness have then led to a surge of interest in defining, evaluating, and improving fairness in ML algorithms. 

The availability of numerous base estimators and \acrfull{bm} methods, however, poses the challenge of selecting the optimal approach for a particular application. Although several comparison studies have been conducted \cite{hufthammer_bias_2020, hort_fairea_2021, chen_comprehensive_2023, adebayo_fairml_2016, hamilton_benchmarking_2017, roth_comparison_2018, friedler_comparative_2018, biswas_machine_2020, chakraborty_fairway_2020}, they did not provide clear best model recommendations. In addition, such studies typically focus only on comparing different \acrshort{bm} methods, leaving out other aspects such as the selection of metrics, base estimators, and classification threshold values.
To address this research gap, this paper proposes
FairGridSearch\footnote{The implementation of the framework is written in Python and available on GitHub: \url{https://github.com/dorisscma/FairGridSearch}} for comparing fairness-enhancing models in binary classification problems.
The framework supports a broad range of parameter-tuning options, including six base estimators, their corresponding parameters, classification thresholds, and nine \acrshort{bm} methods. Furthermore, it provides flexibility in selecting accuracy and fairness metrics for model evaluation. The term "fairness-enhancing models" in this study refers to all models considered in the comparison when taking fairness into account, with or without \acrshort{bm} methods.

\vspace{-3pt}
\section{Related Work and Foundations}
\label{sec:literature_review}
\vspace{-3pt}
\subsection{Related Work}
The growing interest in fairness in \acrshort{ml} has sparked relevant research including definitions, measurements, and \acrshort{bm} methods. 
Some aim to optimize \acrshort{ml} base estimator parameters to achieve desired outcomes \cite{dabra_tune_2023}, while others expand their scope to compare various base estimators, \acrshort{bm} techniques, and metrics \cite{pessach_review_2022}.
\cite{hamilton_benchmarking_2017} compared four fairness metrics and algorithms across three datasets, finding no universally applicable approach; while \cite{roth_comparison_2018} 
identifies \acrfull{lr} as the most versatile base estimator, yielding high accuracy and fairness. 
\cite{friedler_comparative_2018} compared several fairness-aware methods, noting close correlations between group-conditioned metrics; \cite{biswas_machine_2020} evaluated seven \acrshort{bm} methods, stressing post-processing algorithms as the most competitive. \cite{chakraborty_fairway_2020} compared three \acrshort{bm} approaches with their own method Fairway, using exclusively \acrshort{lr} models; \cite{hufthammer_bias_2020} conducted an empirical analysis of \acrfull{roc} and \acrfull{pr} on a binary classification task, showing improved fairness with minimal accuracy cost. 
Fairea \cite{hort_fairea_2021} benchmarks 12 \acrshort{bm} methods and proposes a fairness-accuracy trade-off strategy; \cite{chen_comprehensive_2023} evaluates 12 methods on five datasets, 
making it one of the most comprehensive studies in the literature.

Yet, prior research either has limited evaluation of metrics, \acrshort{bm} methods, and base estimators, or lacks the best model recommendation. FairGridSearch bridges this gap by including several \acrshort{bm} methods, base estimators, and a wide set of accuracy and fairness metrics. 
Most importantly, it recommends the best model considering both accuracy and fairness. Table \ref{table:Comparison_Studies} shows the overview of previous studies and FairGridSearch.


\begin{table}[t]
\caption{Studies Evaluating Different \acrshort{bm} Methods}
\label{table:Comparison_Studies}
\begin{threeparttable}
\centering
\begin{tabularx}{\columnwidth}{YYYYYY}
\toprule
Study                    & \# Datasets & \# BM & \# Acc Metrics  & \# Fair Metrics  & \# Base Est.                     \\
\midrule
\cite{hamilton_benchmarking_2017} & 3 & 3 & 2 & 4 & 2 \\
\cite{roth_comparison_2018}       & 4 & 3 & 3 & 2 & 4 \\
\cite{friedler_comparative_2018}  & 5 & 4 & 4 & 8 & 4 \\
\cite{biswas_machine_2020}        & 5 & 7 & 2 & 7 & varies \\
\cite{chakraborty_fairway_2020}   & 5 & 4 & 2 & 2 & 1 \\
\cite{hufthammer_bias_2020}       & 1 & 2 & 1 & 4 & 2 \\
\cite{hort_fairea_2021}           & 3 & 8 & 1 & 2 & 3 \\
\cite{chen_comprehensive_2023}    & 5 & 17 & 11 & 4 & 4 \\
\textbf{\scriptsize FairGridSearch}    & 3 & 9 & 6 & 8 & 6 \\
\bottomrule
\end{tabularx}

\end{threeparttable}
\end{table}

\vspace{-8pt}
\subsection{Algorithmic Fairness}

\label{sec:fairness_criteria}

Existing fairness criteria fall under two categories: group and individual fairness \cite{zemel_learning_2013}. Group fairness measures statistical parity between different groups based on protected attributes (PA),
while individual fairness requires identical outcomes for similar individuals.
\cite{dwork_fairness_2012}.
Despite myriads of notions to quantify fairness, each measure emphasizes different aspects of what can be considered “fair” \cite{caton_fairness_2020}.
Several studies have shown that it is difficult to satisfy some of the group fairness constraints at once except in highly constrained special cases \cite{mehrabi_survey_2022-1}, or even impossible \cite{berk_fairness_2021, kleinberg_inherent_2016, pleiss_fairness_2017-1}.
In addition, the group fairness criteria generally provide no guarantee for fairness at the individual level either \cite{dwork_fairness_2012}. 
This paper follows "Fairness Tree", an instruction on the selection of fairness criteria by \cite{saleiro_aequitas_2019}.

\vspace{-8pt}
\subsection{Bias Mitigation}
\label{sec:bias_mitigation}
Several attempts have been undertaken to incorporate the concept of fairness into the \acrshort{ml} pipeline, with different approaches and choices of fairness criteria.
These interventions mitigate certain kinds of bias 
at different stages of the \acrshort{ml} pipeline and, depending on the phase they alter, three categories: pre-processing, in-processing, and post-processing can be applied accordingly \cite{barocas_fairness_2019}. 

\vspace{-3pt}
\section{FairGridSearch Framework}
\vspace{-3pt}
\subsection{General Framework}


The FairGridSearch framework resembles conventional GridSearch and allows adjustment of various parameters including base estimators, specific hyper-parameters, classification threshold, and \acrshort{bm} approaches.
Besides, addressing model performance instability, \cite{friedler_comparative_2018} recommend using multiple randomized train-test splits. FairGridSearch incorporates this by executing stratified k-fold cross-validation. The algorithm structure is shown in Algorithm \ref{alg:FairGridSearch}, and subsequent sections provide more comprehensive details on parameter tuning.

\vspace{-5pt}
\begin{algorithm}
\caption{FairGridSearch}\label{alg:FairGridSearch}
\fontsize{8}{12}\selectfont
\SetKwFunction{isOddNumber}{isOddNumber}
\SetKwInOut{KwIn}{Input}
\SetKwInOut{KwOut}{Output}

    \KwIn{dataset $D$, base estimator $base$, parameter grid $param\_grid$, k-fold $k$}
    \KwOut{optimal set of parameters, table of all results}
    \For{hyperp in param\_grid[hyperp\_grid]}{
        \For{train, test in stratified-kfold(D, k)}{
            \For{BM in param\_grid[BM\_grid]}{
                model = BM(base($hyperp$))\;
                model.fit(train)\;
                pred\_prob = model.predict\_proba(test)\;
                \For{threshold in param\_grid[threshold\_grid]}{
                Get prediction with respect to threshold\;
                Calculate accuracy and fairness metrics based on prediction}
                }
            }
         take average of all metrics from k-fold for each model   
         }
    \KwRet{$best\_param$, $result\_table$}
\end{algorithm} 
\vspace{-8pt}

\vspace{-3pt}
\subsection{Parameter Tuning}
\vspace{-3pt}
\subsubsection{Base Estimator}


According to \cite{hort_bias_2022}, the most common classification base estimators in fair \acrshort{ml} are \acrshort{lr} and \acrfull{rf}. Additionally, \cite{hort_bias_2022} found that most publications applied \acrshort{bm} approaches to one base estimator. However, the selection of base estimators can impact model accuracy, and potentially its fairness properties as well. Therefore, six base estimators are included in FairGridSearch, including \acrshort{lr}, \acrshort{rf}, \acrfull{gb}, \acrfull{svm}, \acrfull{nb}, and \acrfull{tabtrans}. 


\subsubsection{Classification Threshold}
Apart from enhancing accuracy, modifying the classification threshold can also play a significant role in model fairness. 
The number of false positives and false negatives varies with the choice of the threshold, so tuning classification thresholds can be utilized as a means of prioritizing between errors, given that the costs of prediction errors may differ as highlighted by \cite{kozodoi_fairness_2022}. 
FairGridSearch enables model optimization by exploring different threshold values.

\subsubsection{Bias Mitigation}
FairGridSearch has nine \acrshort{bm} methods, including two pre-processing methods \acrfull{rw}, \acrfull{lfr}\_pre, three in-processing methods LFR\_in, \acrfull{ad}, \acrfull{egr}, two post-processing methods \acrshort{roc}, \acrfull{ceo}, and two mixed approaches RW+ROC and RW+CEO. Prior research has primarily targeted algorithmic bias by intervening at a single \acrshort{ml} pipeline stage with only one \acrshort{bm} method. Still, bias could persist through other stages \cite{ghai_cascaded_2022}. FairGridSearch addresses this issue by including two mixed approaches.
\acrshort{tabtrans} is incompatible with two \acrshort{bm} approaches. First, LFR\_pre requires numerical conversion of all categorical variables, while \acrshort{tabtrans} requires at least one categorical variable in the dataset; 
and second, \acrshort{egr} provided in AIF360.sklearn is limited to sklearn models, making \acrshort{tabtrans} unfeasible.


\vspace{-3pt}
\subsection{Best Model Criterion}
FairGridSearch selects the optimal model by employing a scoring metric that takes into account both accuracy and fairness metrics.
Following the cost-based analysis method proposed by \cite{haas_price_2019}, the overall cost of a model is determined as a linear combination of accuracy and fairness costs, where respective costs are measured by the distance to the metrics' optimal value. $\alpha$ and $\beta$ are assigned to represent the weights for these costs. The overall cost is hence defined as follows:
\begin{equation}
\label{eq:accfair_cost_analysis}
    C = C_{acc} + C_{fair} = \alpha \cdot (1-metric_{acc}) + \beta \cdot \abs{metric_{fair}}
\end{equation}
and the best model is the one that minimizes overall cost.

\subsubsection{Accuracy Metrics} 
\cite{math11081771} highlight the importance of choosing appropriate accuracy metrics when evaluating fairness-enhancing models, as maximizing one accuracy metric does not guarantee maximization of another. 
FairGridSearch includes several common accuracy metrics such as \acrfull{acc}, \acrfull{bacc}, F1 Score, and AUC. Besides, \acrfull{mcc} is also included in the set of accuracy metrics as suggested by several recent research papers (\cite{canbek_benchmetrics_2021, chicco_advantages_2020, chicco_benefits_2021, chicco_matthews_2021, gosgens_good_2022}).
As \acrshort{mcc} is bounded between -1 and 1, its normalized variant \acrshort{norm_mcc} is included for better comparability to other accuracy metrics, $norm\_MCC = 0.5*(MCC + 1) \in [0,1]$. In total, FairGridSearch comprises six accuracy metrics: \acrshort{acc}, \acrshort{bacc}, F1 score, AUC, \acrshort{mcc}, and \acrshort{norm_mcc}, with \acrshort{norm_mcc} being the primary metric employed in the best model criterion for the exemplary experiments.

\subsubsection{Fairness Metrics} 
\label{sec:method_fair_metrics}
FairGridSearch framework incorporates several group and individual fairness criteria into the algorithm, including \acrfull{spd}, \acrfull{aod}, \acrfull{eod}, \acrfull{ford}, \acrfull{ppvd}, \acrfull{cns}, \acrfull{gei}, and \acrfull{ti}.
The selection of fairness criteria was guided by the "fairness tree" proposed by \cite{saleiro_aequitas_2019}.


\vspace{-3pt}
\section{Exemplary Experiments}
\label{sec:experiment}
The experiments were conducted using all six base estimators and nine \acrshort{bm} methods provided in the framework, and five classification thresholds ranging from 0.3 to 0.7. For each base estimator, four different combinations of base-specific parameters were considered as shown in Table \ref{table:base_parameter}.
Additionally, baseline models without any \acrshort{bm} methods were included for comparison.

\begin{table}[t]
\caption{Base Estimator Parameters}
\label{table:base_parameter}
\centering
\begin{tabularx}{\columnwidth}
{@{\hspace{0.5em}}l@{\hspace{1em}}l@{\qquad}}
\toprule
Base Estimator   &  Parameters  \\ 
\midrule
LR               & {'C':[1, 10], \hspace{4pt}'solver':['liblinear', 'saga']}               \\
RF               & {'n\_estimators':[10, 50], \hspace{4pt}'criterion':['gini', 'entropy']} \\
GB               & {'n\_estimators':[10, 50], \hspace{4pt}'max\_depth':[8, 32]}            \\ 
SVM              & {'kernel':['rbf','linear','poly','sigmoid']}               \\
NB               & {'var\_smoothing': np.logspace(0,-9, num=4)}               \\ 
TabTrans         & {'epochs':[20, 30], \hspace{4pt}'learing\_rate':[1e-04, 1e-05]}         \\
\bottomrule
\end{tabularx}
\end{table}

Three datasets were used in the experiments: Adult, COMPAS, and \acrfull{gc}. These datasets are the most popular ones in the field of algorithmic fairness
\cite{fabris_algorithmic_2022}.
On top of being the most widely used datasets in the relevant research, these three datasets fall into three different categories according to the fairness tree, making them suitable choices for exemplary experiments. 
An overview of all three datasets is shown in Table \ref{table:dataset_overview}.

Table \ref{table:n_models_in_experiment} shows the number of models for each dataset. \acrshort{tabtrans} models were run with two fewer bias mitigators since they are incompatible with LFR\_pre and EGR. The third row shows the two base estimator invariant \acrshort{bm} methods, LFR\_in and AD. These two in-processing methods change the entire model algorithm and therefore do not take base estimators into account. In total, 930 models were implemented for each dataset and each model was run with 10-fold cross-validation.

For the best model criterion, \acrshort{norm_mcc} was chosen as the accuracy metric across all datasets, while the selection of fairness metric was guided by the fairness tree from \cite{saleiro_aequitas_2019}. 
\acrshort{spd} was chosen as the fairness metric for the Adult dataset, as there was originally no prediction-based intervention intended. In the COMPAS setting, where predictions are used for pretrial release decisions, \acrshort{ppvd} was chosen as the fairness metric, considering it as a punitive intervention. Lastly, as issuing credit is regarded as an assistive intervention, \acrshort{eod} is chosen as the fairness metric for the German Credit dataset.
For all fairness metrics, the absolute values were used to indicate the magnitude of bias, regardless of the direction. Values near zero indicate less bias, while values further away indicate more bias. Finally, the weights of the accuracy and fairness metrics
were both set to 1, meaning equal consideration for both criteria. 

\begin{table}[h]
\caption{Number of Models in the Experiments}
\label{table:n_models_in_experiment}
\centering
\begin{tabularx}{\columnwidth}{l *5{>{\centering\arraybackslash}X}}
\toprule
                 & Base & Param. & $\tau$ & BM & Total \\ 
\midrule
All but TabTrans   & 5    & 4       & 5         & 8  & 800   \\
TabTrans          & 1    & 4       & 5         & 6  & 120   \\
Base-invariant BMs & -    & -       & 5         & 2  & 10    \\ 
\bottomrule
\end{tabularx}
\end{table}

The experiments were conducted on the CPU instance of SageMaker Studio Lab
with RAM of 16 GB, except for SVM models on the Adult dataset, which were run on an M1 Pro chip with a CPU speed of 3228 MHz and 16 GB of RAM. For SageMaker Studio Lab, the CPU speed information was not available due to the availability of compute instances being subject to demand. The completion times for all models varied across the datasets. The Adult dataset took the longest to process, totaling 221 hours; COMPAS 4.7 hours, and \acrshort{gc} only 2.1 hours.

\begin{table}[t]
\centering
\caption{Datasets Used in the Experiments}
\label{table:dataset_overview}
\begin{tabularx}{\columnwidth}
{@{\hspace{0.5em}}l@{\hspace{0.5em}}c@{\hspace{0.5em}}c@{\hspace{0.5em}}c@{\hspace{0.5em}}c@{\hspace{0.5em}}c@{\hspace{0.em}}}
\toprule
              & Shape           & PA (priv.)           & Fav. Label &  Fair Metric \\
\midrule
Adult         & (46,447, 14)    & Race (Caucasian)    & High Income     & SPD \\
COMPAS        & (6,150, 9)      & Race (White)        & No Recidivism   & PPVD \\
\acrshort{gc} & (1,000, 21)     & Sex  (Male)         & Good Credit     & EOD \\ 
\bottomrule
\end{tabularx}
\end{table}


\vspace{-3pt}
\section {Results}

\vspace{-3pt}
\subsection{Top Models}
FairGridSearch evaluates different fairness-enhancing models across various combinations of model parameters. The model with the lowest accuracy and fairness cost is then selected as the top model.
The top models for all three datasets are shown in Table \ref{table:top_models}. 
\begin{table*}[h]
\caption{Top Models for All Three Datasets}
\label{table:top_models}
\begin{tabularx}{\textwidth}{lcXcccccc}
\toprule
Dataset         & Base & Param. & BM & $\tau$ & Norm. MCC & Abs. Fair & Cost   \\ 
\midrule
Adult           & GB             & \RaggedRight{{'criterion': 'friedman\_mse', 'max\_depth': 8, 'n\_estimators': 50}} & RW+ROC & 0.4 & 0.8115 & 0.0091 (SPD) & 0.1976 \\
Compas          & GB             & \RaggedRight{{'criterion': 'friedman\_mse', 'max\_depth': 8, 'n\_estimators': 10}} & RW+ROC & 0.5 & 0.6506 & 0.0019 (PPVD) & 0.3512 \\
\acrshort{gc}   & RF             & {'criterion': 'gini', 'max\_depth': 16, 'n\_estimators': 50} & EGR & 0.7 & 0.7062 & 0.0003 (EOD) & 0.2941 \\
\bottomrule
\end{tabularx}
\end{table*}

\vspace{-3pt}
\subsection{Metrics}
\vspace{-3pt}
\subsubsection{Correlation between Metrics}
To better understand the relationship between metrics, the following sections illustrate Spearman's rank correlation coefficient $\rho$ between every pair of the metrics with heatmaps.
A positive $\rho$ indicates a tendency for both variables to increase, while a negative $\rho$ suggests an inverse relationship.
For each metric pair, the overall $\rho$ is presented along with stars indicating the statistical significance level  (i.e., *, **, *** indicating p-value < 0.05, 0.01, 0.001, respectively).

\begin{figure*}[t]
\centering
\begin{subfigure}{0.28\textwidth}
    \includegraphics[width=\textwidth]{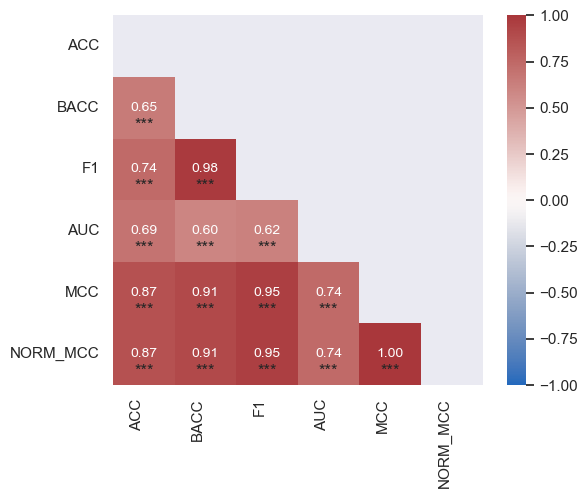}
    \caption{Adult}
    \label{fig:Adult_original_acc_corr}
\end{subfigure}
\hspace{12pt}
\begin{subfigure}{0.28\textwidth}
    \includegraphics[width=\textwidth]{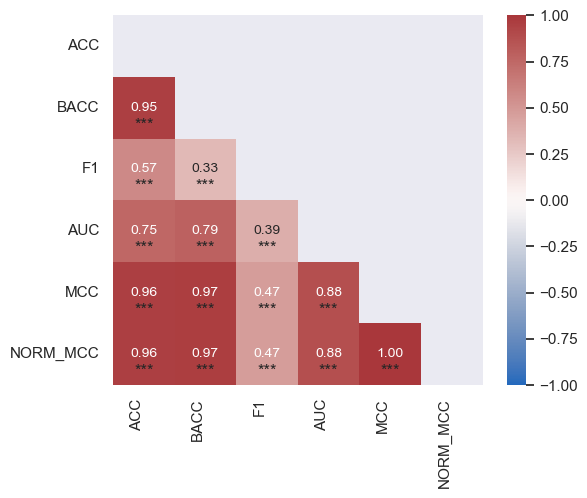}  
    \caption{COMPAS}
    \label{fig:Compas_original_acc_corr}
\end{subfigure}
\hspace{12pt}
\begin{subfigure}{0.28\textwidth}
    \includegraphics[width=\textwidth]{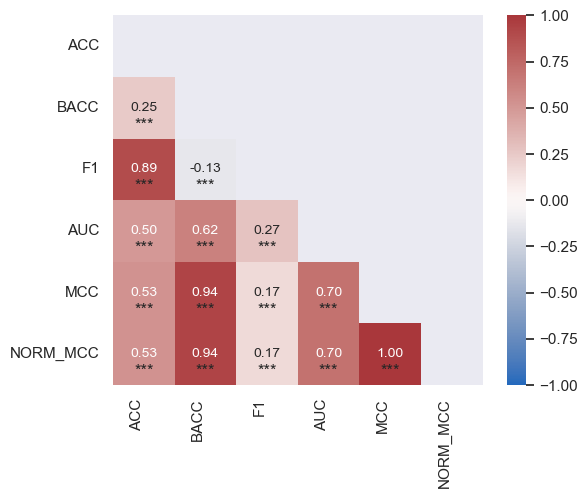} 
    \caption{German Credit}
    \label{fig:German_Credit_original_acc_corr}
\end{subfigure}
\caption{(NORM\_MCC, BACC) is the only accuracy metric pair showing high positive correlations across all datasets.}
\label{fig:original_acc_corr}
\end{figure*}

\begin{figure*}[t]
\centering
\begin{subfigure}{0.27\textwidth}
    \includegraphics[width=\textwidth]{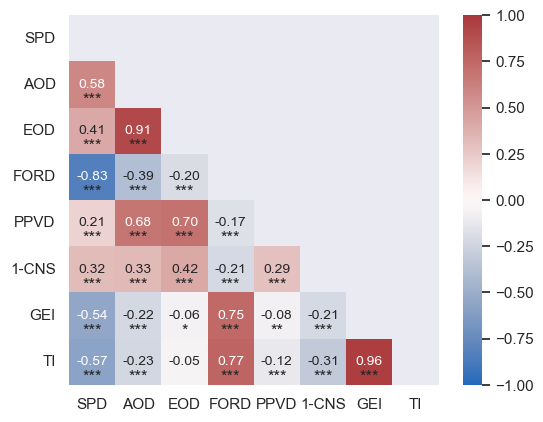}
    \caption{Adult}
    \label{fig:Adult_original_fair_corr}
\end{subfigure}
\hspace{12pt}
\begin{subfigure}{0.27\textwidth}
    \includegraphics[width=\textwidth]{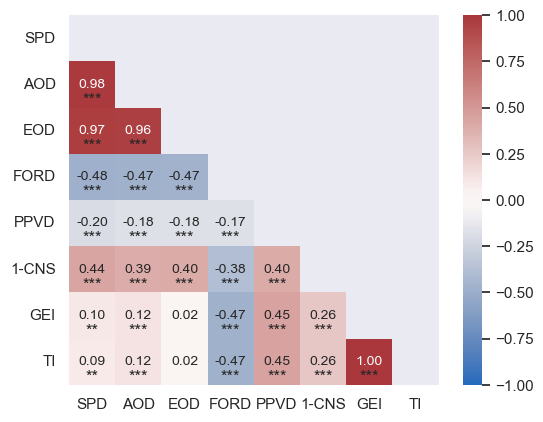}  
    \caption{COMPAS}
    \label{fig:Compas_original_fair_corr}
\end{subfigure}
\hspace{12pt}
\begin{subfigure}{0.27\textwidth}
    \includegraphics[width=\textwidth]{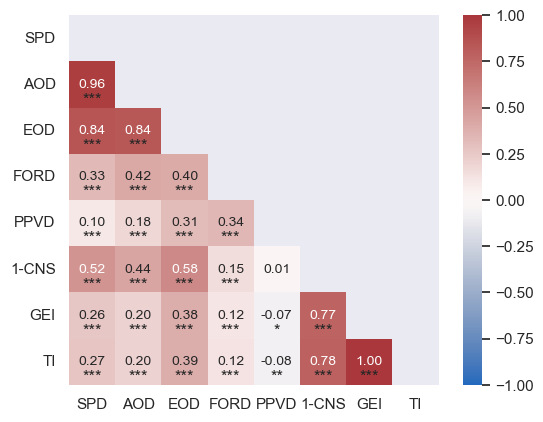}
    \caption{German Credit}
    \label{fig:German_Credit_original_fair_corr}
\end{subfigure}
\caption{Correlation between fairness metrics varies substantially across datasets.}
\label{fig:original_fair_corr}
\vspace{-12pt}
\end{figure*}

Fig. \ref{fig:original_acc_corr} illustrates Spearman's rank correlation coefficient $\rho$ between accuracy metrics. 
Almost all metric pairs exhibit a positive correlation, implying that accuracy metrics tend to move in the same direction.
However, the degree of correlation differs considerably across the datasets. 
The only exception to this trend is the (MCC/NORM\_MCC, BACC) pair, which consistently shows a high positive correlation across all datasets.
Fig. \ref{fig:original_fair_corr} reveals that the correlation between fairness metrics also varies substantially across datasets.
The metrics of SPD, EOD, and AOD generally exhibit higher correlations with each other, particularly in the COMPAS and German Credit datasets. 
Furthermore, the FORD metric shows a negative correlation with other fairness metrics in Adult and COMPAS datasets, indicating that when FORD increases, the other fairness metrics tend to decrease. 
This negative correlation is especially high for the (FORD, SPD) metric pair in the Adult dataset. 
\subsubsection{Metric Changes after BM}
\label{sec:results_cross_dataset}
To further investigate the importance of metric selection, this section employs the methodology proposed by \cite{chen_comprehensive_2023} and examines responses from different metrics to \acrshort{bm} methods. 
First, a non-parametric Mann-Whitney U-test determines the statistical significance of differences between the same models before and after applying \acrshort{bm}, with a significance level of 0.05. Second, Cohen's $d$ effect size assesses whether the difference has a substantive effect.  
According to \cite{chen_comprehensive_2023}, effect sizes 
$d \in [0, 0.5)$ are considered small, $d \in [0.5, 0.8)$ medium, and $d \in [0.8, \infty)$ large. 

\begin{figure*}[t]
\centering
\begin{subfigure}{0.24\textwidth}
    \includegraphics[width=\textwidth]{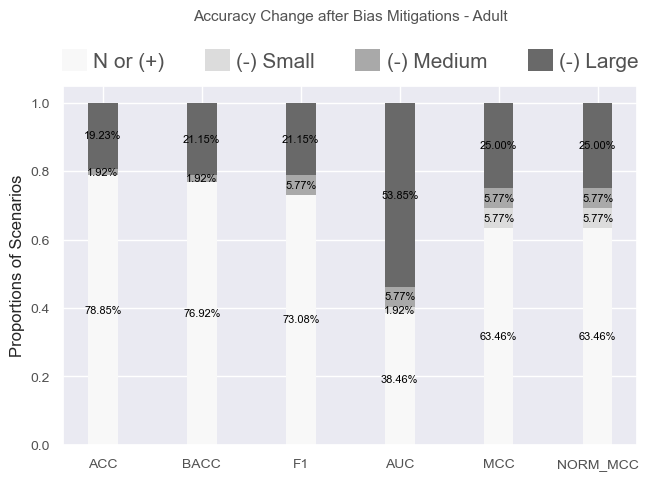}
    \caption{Adult}
    \label{fig:acc_change_metrics_adult}
\end{subfigure}
\begin{subfigure}{0.24\textwidth}
    \includegraphics[width=\textwidth]{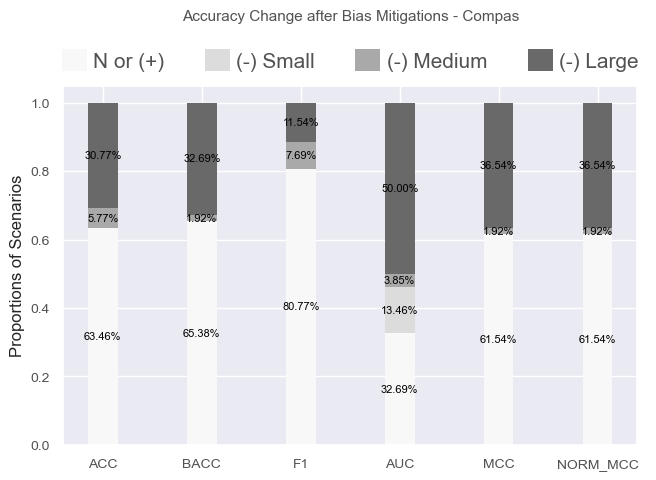}  
    \caption{COMPAS}
    \label{fig:acc_change_metrics_compas}
\end{subfigure}
\begin{subfigure}{0.24\textwidth}
    \includegraphics[width=\textwidth]{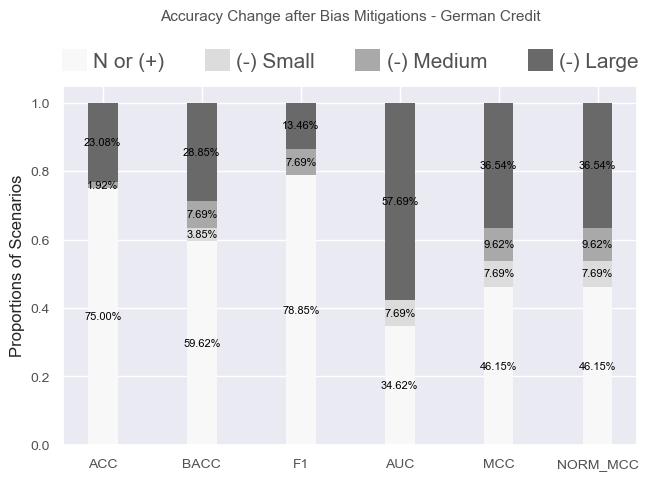} 
    \caption{German Credit}
    \label{fig:acc_change_metrics_gc}
\end{subfigure}
\begin{subfigure}{0.24\textwidth}
    \includegraphics[width=\textwidth]{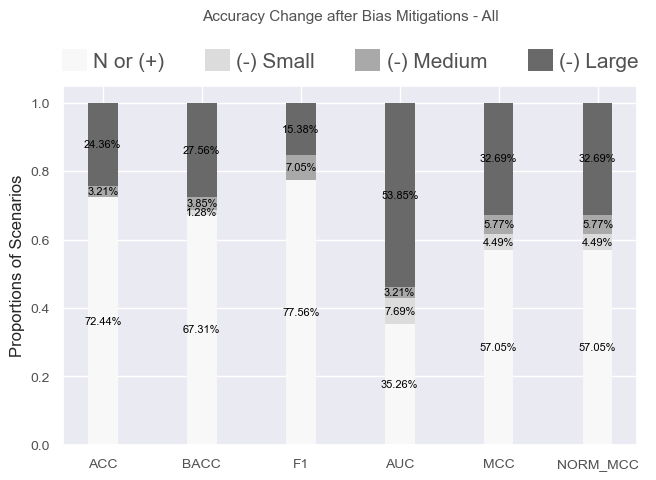} 
    \caption{All}
    \label{fig:acc_change_metrics_all}
\end{subfigure}
\caption{Accuracy metrics respond differently to bias mitigators.}
\label{fig:acc_change_metrics}
\end{figure*}

\begin{figure*}[t]
\centering
\begin{subfigure}{0.24\textwidth}
    \includegraphics[width=\textwidth]{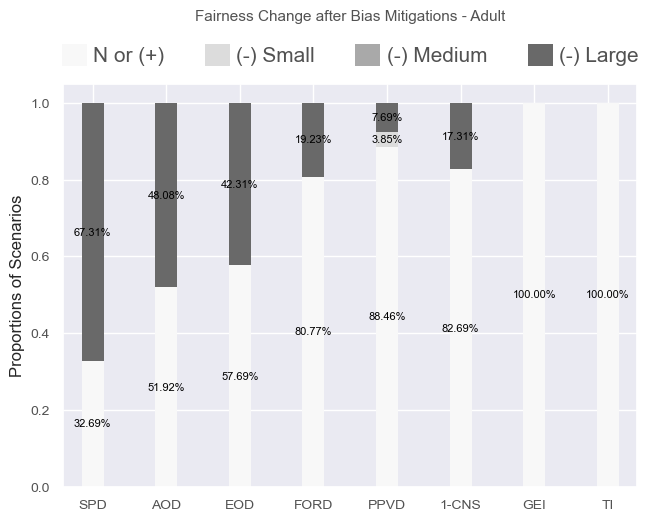}
    \caption{Adult}
    \label{fig:fair_change_metrics_adult}
\end{subfigure}
\begin{subfigure}{0.24\textwidth}
    \includegraphics[width=\textwidth]{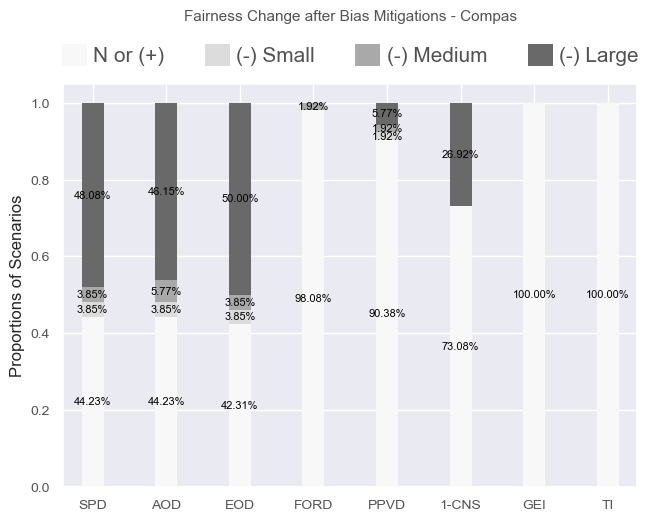}  
    \caption{COMPAS}
    \label{fig:fair_change_metrics_compas}
\end{subfigure}
\begin{subfigure}{0.24\textwidth}
    \includegraphics[width=\textwidth]{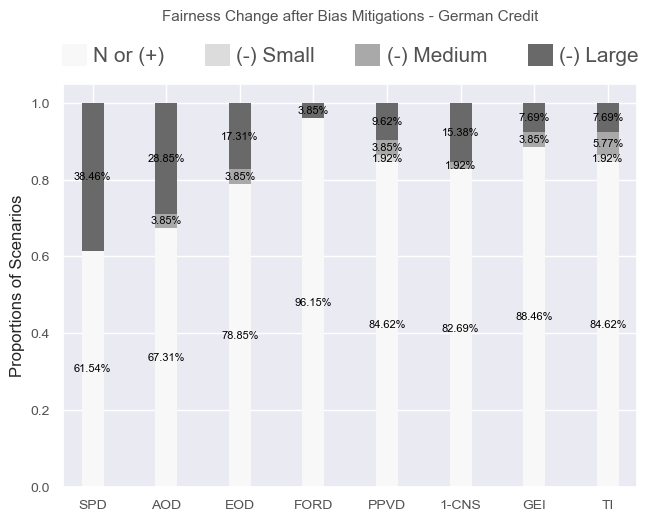} 
    \caption{German Credit}
    \label{fig:fair_change_metrics_gc}
\end{subfigure}
\begin{subfigure}{0.24\textwidth}
    \includegraphics[width=\textwidth]{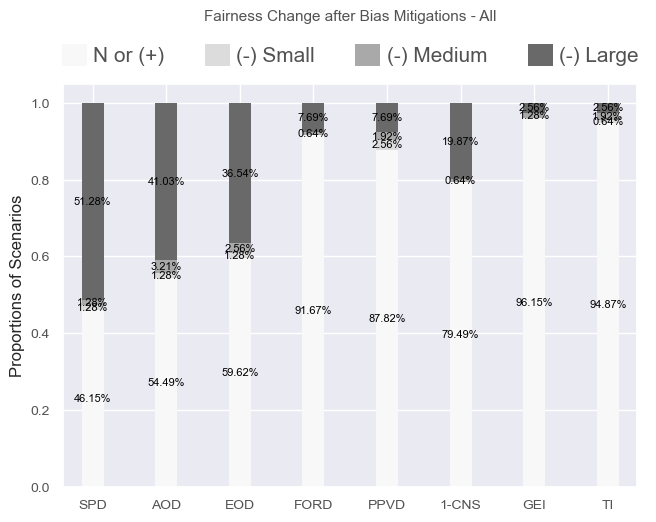} 
    \caption{All}
    \label{fig:fair_change_metrics_all}
\end{subfigure}
\caption{Efficacy of \acrshort{bm} methods varies across different fairness metrics.}
\label{fig:fair_change_metrics}
\end{figure*}

Fig. \ref{fig:acc_change_metrics_all} reveals that the accuracy metrics decrease significantly in an average of 40\% of all scenarios (ranging from 22.44\% to 64.74\% depending on the metric). This suggests that different accuracy metrics exhibit varying degrees of sensitivity to \acrshort{bm} application. AUC, for instance, drops the most among all metrics across all datasets.
Fig. \ref{fig:fair_change_metrics} illustrate varying \acrshort{bm} efficacy across fairness metrics. Here, since the optimal value for fairness metrics is zero, a decrease in fairness metrics indicates a reduction in bias. Notably, individual fairness metrics show minimal improvement after \acrshort{bm} implementations, this is anticipated as the methods are primarily designed to improve group fairness, and enhancing group fairness does not guarantee improvement in individual fairness \cite{dwork_fairness_2012}.
However, even certain group fairness criteria, such as FORD and PPVD, display limited effectiveness, particularly in the COMPAS dataset.
Overall, metrics are not always correlated with their alternatives and exhibit distinct reactions to \acrshort{bm} methods. This holds true for both accuracy and fairness metrics, underscoring the need for careful metric selection when evaluating models.

\vspace{-3pt}
\subsection{Base Estimator}
\begin{figure*}[ht]
\centering
\begin{subfigure}{0.24\textwidth}
    \includegraphics[width=\textwidth]{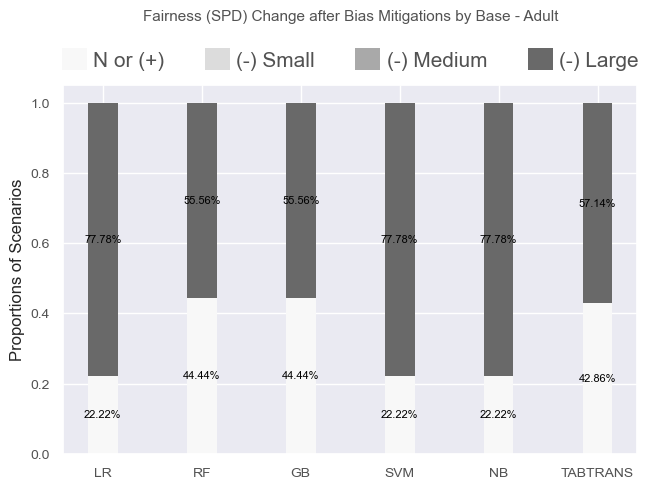}
    \caption{Adult}
    \label{fig:fair_change_base_adult}
\end{subfigure}
\begin{subfigure}{0.24\textwidth}
    \includegraphics[width=\textwidth]{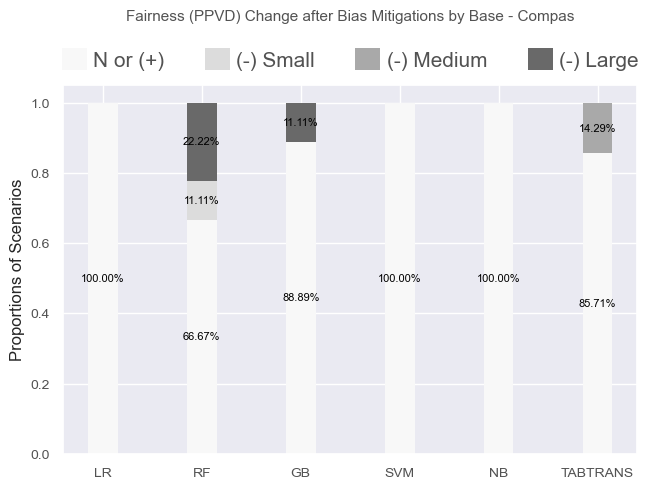}  
    \caption{COMPAS}
    \label{fig:fair_change_base_compas}
\end{subfigure}
\begin{subfigure}{0.24\textwidth}
    \includegraphics[width=\textwidth]{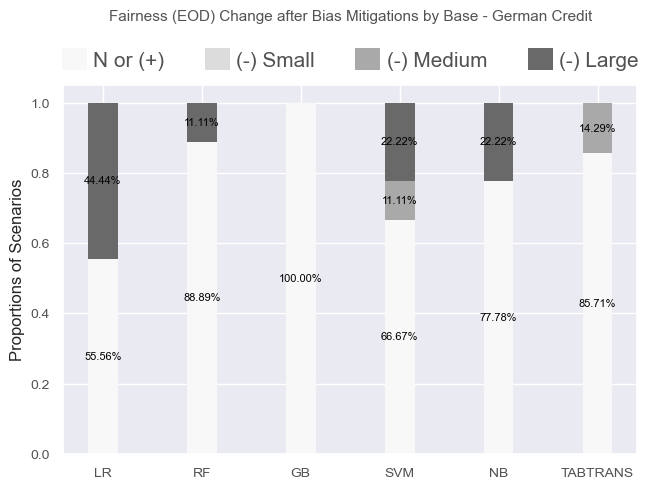} 
    \caption{German Credit}
    \label{fig:fair_change_base_gc}
\end{subfigure}
\begin{subfigure}{0.24\textwidth}
    \includegraphics[width=\textwidth]{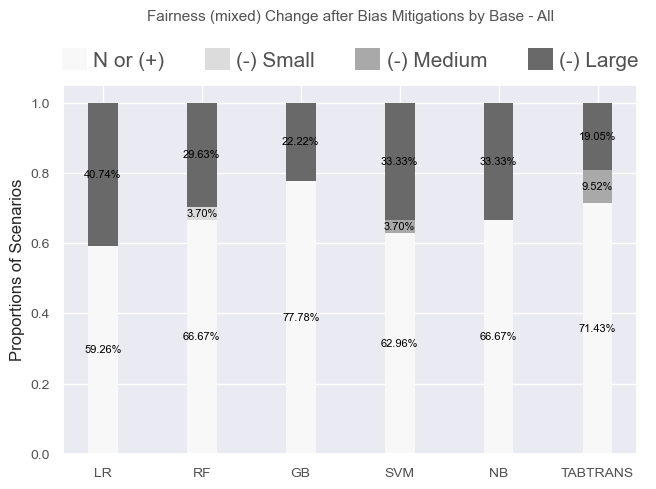} 
    \caption{All}
    \label{fig:fair_change_base_all}
\end{subfigure}
\caption{No single base estimator consistently outperforms the others.}
\label{fig:fair_change_base}
\vspace{-12pt}
\end{figure*}

    From Fig. \ref{fig:fair_change_base}, it can be observed that no base estimator consistently improve model fairness better than the others across all datasets. Base estimators exhibit varying degrees of sensitivity to \acrshort{bm} in different datasets. 
    For example, \acrshort{lr} models applied with \acrshort{bm} methods improved fairness in the Adult and German Credit datasets but not in the COMPAS dataset. Similarly, \acrshort{gb} models enhance fairness in the Adult and COMPAS datasets but not in the German Credit datasets. These findings highlight the absence of a universal solution for selecting the optimal base estimator for fair models.
    
\vspace{-3pt}
\subsection{Classification Threshold}
\begin{figure*}
\label{fig:2D_threshold}
\centering
\begin{subfigure}{0.29\textwidth}
    \includegraphics[width=\textwidth]{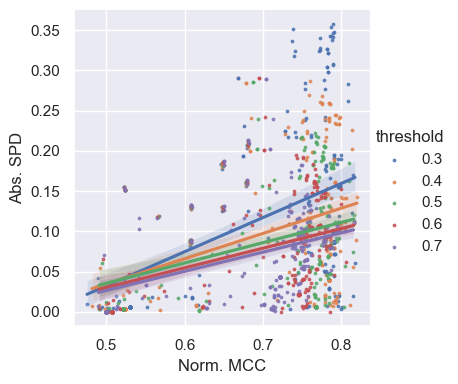} 
    \caption{Adult}
    \label{fig:Adult_threshold}
\end{subfigure}
\hspace{3pt}
\begin{subfigure}{0.29\textwidth}
    \includegraphics[width=\textwidth]{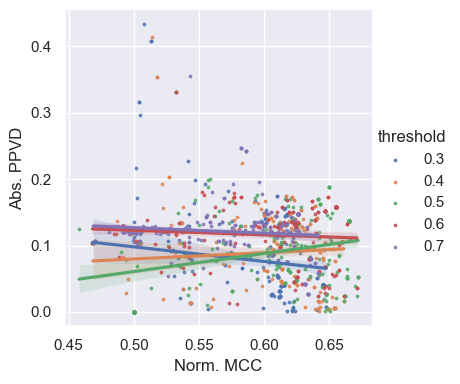} 
    \caption{COMPAS}
    \label{fig:COMPAS_threshold}
\end{subfigure}
\hspace{3pt}
\begin{subfigure}{0.29\textwidth}
    \includegraphics[width=\textwidth]{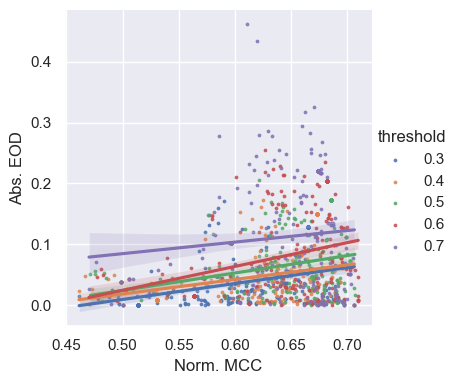} 
    \caption{German Credit}
    \label{fig:german_credit_threshold}
\end{subfigure}
\caption{Optimal threshold value varies with datasets.}
\end{figure*}

  Fig. \ref{fig:Adult_threshold} and \ref{fig:german_credit_threshold} show that accuracy-maximizing threshold values may also generate high volatility in fairness. In contrast, threshold values that have greater stability in fairness tend to limit the model accuracy. 
  Given the variability in optimal classification threshold across datasets and the significant impact it has, it is advisable to include a comprehensive set in the comparison to identify the most suitable one.

\vspace{-3pt}
\subsection{Bias Mitigation}
\begin{figure*}[ht]
\centering
\begin{subfigure}{0.24\textwidth}
    \includegraphics[width=\textwidth]{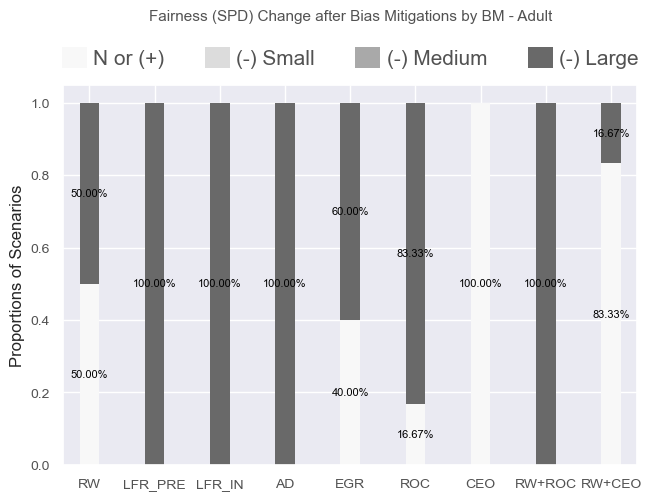}
    \caption{Adult}
    \label{fig:fair_change_bm_adult}
\end{subfigure}
\begin{subfigure}{0.24\textwidth}
    \includegraphics[width=\textwidth]{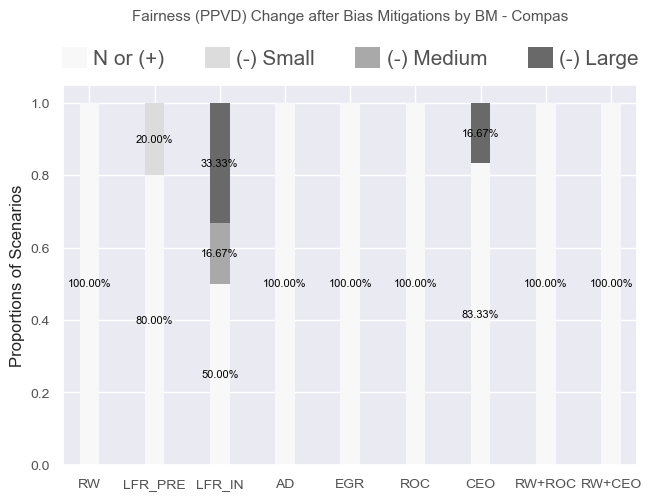}  
    \caption{COMPAS}
    \label{fig:fair_change_bm_compas}
\end{subfigure}
\begin{subfigure}{0.24\textwidth}
    \includegraphics[width=\textwidth]{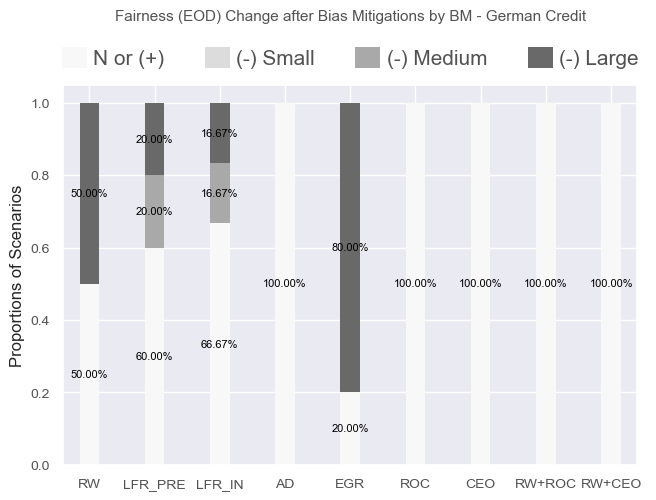} 
    \caption{German Credit}
    \label{fig:fair_change_bm_gc}
\end{subfigure}
\begin{subfigure}{0.24\textwidth}
    \includegraphics[width=\textwidth]{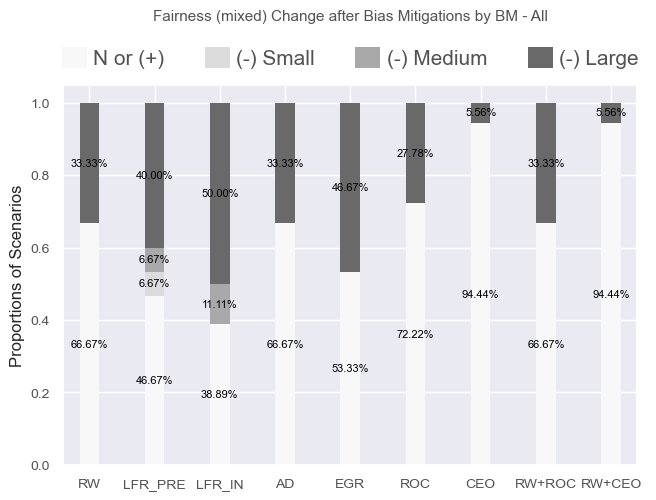} 
    \caption{All}
    \label{fig:fair_change_bm_all}
\end{subfigure}
\caption{No single \acrshort{bm} method consistently outperforms the others.}
\label{fig:fair_change_bm}
\vspace{-12pt}
\end{figure*}

The results depicted in Fig. \ref{fig:fair_change_bm} reveal substantial variations in the effectiveness of \acrshort{bm} methods, with inconsistent effects across datasets. Effective methods in one dataset may yield no discernible improvements in others. For instance, the ROC method improves model fairness in most cases for the Adult dataset but exhibits no significant effects for the COMPAS and German Credit datasets.
Moreover, mixed approaches do not offer greater enhancement of fairness. In fact, they are even less effective than their constituent individual methods.

\vspace{-16pt}
\section{Discussion and Limitation}
\label{sec:discussion}
The literature has recently seen a multitude of proposals for fairness metrics and bias mitigation methods. However, studies comparing different methods often have limited scope, focusing only on specific model combinations and datasets lacking model recommendations.
To address this gap, the FairGridSearch approach offers two main advantages: facilitating easy implementation and comparison of fairness-enhancing models, and identifying the most suitable one for a given dataset.
    Furthermore, several key findings emerge in our exemplary experiments using three popular datasets. Firstly, the selection of both accuracy and fairness metrics plays a crucial role in model evaluation, given their lack of consistent correlation and distinct responses to \acrshort{bm} methods across datasets. 
    This differs from the findings in \cite{friedler_comparative_2018}, where they claimed strong correlations among fairness metrics. The disparity in results could be attributed to the differences in datasets and model configuration.
    Secondly, no single base estimator consistently outperforms others in improving model fairness when utilized with bias mitigators.
    Thirdly, the choice of classification threshold values can introduce varying degrees of volatility in model fairness. High volatility thresholds may achieve both high fairness and accuracy, while more stable fairness thresholds tend to exhibit lower accuracy. Lastly, the effectiveness of \acrshort{bm} methods is contingent on the dataset, with no single method outperforming others consistently across all datasets.
    Overall, these findings highlight the importance of selecting appropriate metrics and considering multiple factors when building fair \acrshort{ml} models, such as base estimators and classification threshold values, in addition to \acrshort{bm} methods.


This study also comes with some limitations. First, the framework is designed specifically for binary classification problems and does not currently extend to multi-class classification. Second, although the framework includes several commonly used \acrshort{bm} methods, there are numerous other methods available in the literature still to be considered. Thirdly, while grid search is effective for parameter tuning, its computational requirements can be significant. Exploring alternative optimization methods could offer more efficient solutions.
Moreover, the exemplary experiments are conducted on three specific datasets, limiting current generalizability. Including more datasets is hence needed, particularly in light of the differences between our work and \cite{friedler_comparative_2018}.
Future plans thus include expanding the framework with more \acrshort{bm} methods and datasets, as well as including alternative parameter optimization methods.

\vspace{-5pt}
\section{Conclusion}
\label{sec:conclusion}
Despite the rapid growth in the field of fairness in \acrshort{ml}, selecting the optimal fairness-enhancing model remains challenging.
This study focuses on binary classification and proposes the FairGridSearch framework, which facilitates the implementation and comparison of various fairness-enhancing models and suggests the most suitable model for a given application. 
Our experiments emphasize that there's no silver bullet  when building fair \acrshort{ml} models since metric selections, base estimators, classification thresholds, and BM methods all play important roles, underscoring the importance of considering all these factors. By leveraging FairGridSearch, researchers and practitioners can effectively determine the best fairness-enhancing model for their needs.


\vspace{-5pt}
\bibliographystyle{IEEEtran}
{\scriptsize\bibliography{main.bib}}


\end{document}